\documentclass[]{llncs}
\usepackage{times}
\usepackage{latexsym}
\usepackage{booktabs}
\usepackage{multirow}
\usepackage{graphicx}
\usepackage{caption}
\usepackage{subcaption}
\usepackage{float}
\usepackage{lipsum}  
\usepackage{amsfonts,amsmath}

\usepackage{array}
\newcolumntype{L}[1]{>{\raggedright\let\newline\\\arraybackslash\hspace{0pt}}m{#1}}
\newcolumntype{C}[1]{>{\centering\let\newline\\\arraybackslash\hspace{0pt}}m{#1}}
\newcolumntype{R}[1]{>{\raggedleft\let\newline\\\arraybackslash\hspace{0pt}}m{#1}}

\newcommand\blfootnote[1]{%
  \begingroup
  \renewcommand\thefootnote{}\footnote{#1}%
  \addtocounter{footnote}{-1}%
  \endgroup
}

\usepackage{microtype}
\usepackage[normalem]{ulem}
\useunder{\uline}{\ul}{}

\newcommand\precision{\textit{precision}}
\newcommand\recall{\textit{recall}}
\newcommand\fallout{\textit{fallout}}
\newcommand\exmodel{$\mathcal{E}$}
\newcommand\mlmodel{$\mathcal{M}$}

\title{On the Evaluation of the Plausibility and Faithfulness of Sentiment Analysis Explanations}
\author{Julia El Zini \and
Mohamad Mansour \and
Basel Mousi \and
Mariette Awad
}

\authorrunning{J. El Zini et al.}
\institute{Electrical and Computer Engineering Department\\American University of Beirut, Beirut, Lebanon\\
\email{\{jwe04,mgm35,bam20\}@mail.aub.edu, mariette.awad@aub.edu.lb}\\
}

\begin{document}
\maketitle
\begin{abstract}
\blfootnote{The final publication is available at Springer via \url{https://doi.org/10.1007/978-3-031-08337-2\_28}. Citation: El Zini, J., Mansour, M., Mousi, B., Awad, M. (2022). On the Evaluation of the Plausibility and Faithfulness of Sentiment Analysis Explanations. In Artificial Intelligence Applications and Innovations. AIAI 2022. IFIP Advances in Information and Communication Technology, vol 647. Springer, Cham.}
With the pervasive use of Sentiment Analysis (SA) models in financial and social settings, performance is no longer the sole concern for reliable and accountable deployment. SA models are expected to explain their behavior and highlight textual evidence of their predictions. Recently, Explainable AI (ExAI) is enabling the ``third AI wave'' by providing explanations for the highly non-linear black-box deep AI models. Nonetheless, current ExAI methods, especially in the NLP field, are conducted on various datasets by employing different metrics to evaluate several aspects. The lack of a common evaluation framework is hindering the progress tracking of such methods and their wider adoption. 

In this work, inspired by offline information retrieval, we propose different metrics and techniques to evaluate the explainability of SA models from two angles. First, we evaluate the strength of the extracted ``rationales'' in \textit{faithfully} explaining the predicted outcome. Second, we measure the agreement between ExAI methods and human judgment on a homegrown dataset\footnote{Dataset and code available at \url{https://gitlab.com/awadailab/exai-nlp-eval}} to reflect on the rationales \textit{plausibility}. Our conducted experiments comprise four dimensions: (1) the underlying architectures of SA models, (2) the approach followed by the ExAI method, (3) the reasoning difficulty, and (4) the homogeneity of the ground-truth rationales.

We empirically demonstrate that \textit{anchors} explanations are more aligned with the human judgment and can be more confident in extracting supporting rationales. As can be foreseen, the reasoning complexity of sentiment is shown to thwart ExAI methods from extracting supporting evidence. Moreover, a remarkable discrepancy is discerned between the results of different explainability methods on the various architectures suggesting the need for consolidation to observe enhanced performance. Predominantly, transformers are shown to exhibit better explainability than convolutional and recurrent architectures. Our work paves the way towards designing more interpretable NLP models and enabling a common evaluation ground for their relative strengths and robustness.
\end{abstract}

\section{Introduction}\label{sec:intro}
Sentiment Analysis (SA) is instrumental to the financial services industry \cite{rambocas2013marketing,rambocas2018online} as it develops techniques to interpret customer feedback, monitor product reputations, understand the customers' needs, and conduct market research. Harnessing the power of Deep Learning (DL) in understanding general contexts, the performance of SA models is considerably boosted \cite{hassan2017deep,habimana2020sentiment,onan2021sentiment}. However, the non-linearity and the black-box nature of such models hinder the interpretation of the predictions \cite{Mishra2022,samek2019towards}. Besides providing guarantees on reliability, generalization, robustness, and fairness, the interpretability of the SA models can be of service to behavioral marketing and personalized advertisement.

Recently, \textbf{Ex}plainable \textbf{A}rtificial \textbf{I}ntelligence (ExAI) algorithms are breathing a new flexibility in general AI applications by developing methods to explain model's prediction \cite{Chattopadhay_2018,zeilerF13,lime}. Numerical data frameworks and computer vision applications have witnessed an explosive growth of ExAI nurtured by the ease of expression of features as interpretable components \cite{qin2018convolutional,nguyen2019understanding,guidotti2018survey}. However, only a few ExAI methods are applied to textual classifiers, embeddings, and language models \cite{danilevsky2020survey}. In the SA framework, researchers integrated data augmentation techniques to improve the interpretability of SA models \cite{chen2019improving}, studied attention mechanisms in SA through an explainability lens \cite{bodria2020explainability} and applied ExAI on aspect-based SA models \cite{silveira2019using}. 
To date, ExAI methods on Natural Language Processing (NLP) tasks are not evaluated on standardized benchmarking datasets through common metrics which hinders the progress and adoption of such methods in the NLP field. Evaluating explainability methods is two-fold. First, it helps assess the extent to which a deep model can be made explainable. Second, it provides a common ground to measure the contrast between explanations produced by diverse ExAI approaches. 

In this work, we inspect two human aspects of explainability methods: (1) faithfulness to the model being explained and (2) plausibility from a human lens. For this purpose, we select eight state-of-the-art SA models with underlying architectures of recurrent, convolutional, and attention layers. We generate explanations of the predictions of these models on three ExAI methods that can be applied in NLP; mainly LIME \cite{lime}, \textit{anchors} \cite{ribeiro2018anchors} and SHAP \cite{shap}. The generated explanations are then evaluated through two procedures. First, \textit{faithfulness}\footnote{refers to the metric hereafter} is evaluated by examining the degradation in the model's performance when only extracted rationales are fed to the model. Second, the \textit{plausibility}$^1$ of extracted rationales is evaluated via comparison to the human judgment of what a sufficient explanation is. This experiment entails a homegrown dataset of manually labeled explanations on SA data aggregated through conjunction and disjunction means. The comparison is achieved on six proposed metrics, inspired by information retrieval, to evaluate the precision and fallout of exAI methods on the SA models. Hence, our evaluation is carried out over four different dimensions: (1) SA model, (2) ExAI method, (3) reasoning complexity, and (4) human judgment homogeneity. 

The contributions of this work are: (1) a dataset for SA explainability labeled on different dimensions (2) the first faithfulness and plausibility evaluation inspired from information retrieval (3) a thorough four-dimensional ExAI evaluation on SA models.

Our empirical analysis allows us to draw conclusions on the faithfulness of LIME rationales and the plausibility of the anchors model which is found to be more confident in extracting supporting evidence. Moreover, we highlight the consistency of different attention architectures in deriving relatively more plausible explanations. 

Next, we provide a general background on the sentiment analysis and ExAI models used in this work in Section~\ref{sec:background}. Then, we present our evaluation dataset and framework in Sections \ref{sec:dataset} and \ref{sec:method}. We report our comparative analysis in Section~\ref{sec:results} before concluding with final remarks in Section \ref{sec:conc}.

\section{Background}\label{sec:background}
Little has been done on the evaluation of ExAI in NLP settings. Recently, a framework to evaluate rationalized explanations is introduced in ERASER \cite{deyoung2020eraser}. ERASER provides benchmarking data for 7 NLP tasks and suggests sufficiency and comprehensiveness as evaluation metrics. While ERASER considers a wider range of NLP tasks; it is narrow in terms of the deep architectures and angles that it considers. In contrast, we study the explainability of SA from four different perspectives. Prior to ERASER, the work of \cite{ArrasHMMS16a} evaluates the explainability of SVMs compared to CNNs to find that the latter models yield more interpretable decisions. Other attempts only consider the attention mechanism and the debate concerning its inherent interpretability \cite{serrano2019attention,mullenbach2018explainable,jain2019attention}. Additionally, the concept of explanation faithfulness has been introduced before \cite{jacovi2020towards,lime} with no explicit evaluation of textual classifiers.

Next, we provide the background on sentiment analysis and explainability methods. 

    \subsection{Sentiment Analysis Models}
    SA models currently exploit deep architectures, word embeddings, transfer learning, and attention mechanisms \cite{shi2019survey}. In this work, we experiment with eight state-of-the-art SA models with different architectures that vary between convolution, recurrent, and attention networks. The models are chosen with a diversity of word embeddings and some of the models leverage transfer learning techniques during training.

First, \textit{CNN-MC}, a CNN for sentiment classification \cite{cnnmc}, consists of a simple CNN trained on top of static \textit{word2vec} word embeddings. 
To experiment with different embedding models, the Universal Sentence Encoder (USE)\cite{use}, is used to train additional convolution layers. USE obtains sentence embeddings through a deep averaging network. 
The recurrent architecture is tested on the Byte-multiplicative LSTMs (\textit{bmLSTM}),  used in \cite{byte_multiplicative} to generate reviews and discover sentiments.

The attention architecture \cite{vaswani2017attention} is studied in five transformer models. We first study the parent transformer, \textit{BERT} \cite{devlin2018bert} which pre-trains deep bidirectional representations by jointly conditioning on the left as well as the right context extracted by each layer on unlabeled data. Then, we study an optimized version of BERT, \textit{RoBERTa} \cite{liu2019roberta}, which is trained longer and on prolonged sequences. 
We further study a different optimization of BERT, \textit{ALBERT} \cite{lan2020albert}, that factorizes the embedding matrix into smaller matrices, separating thus the size of vocabulary from the size of the hidden layers.
Moreover, we consider \textit{DistilBERT} \cite{sanh2020distilbert}, a distilled version of BERT trained on very large batches while leveraging the computation of the gradients through dynamic masking. 
Finally, a non-BERT transformer, \textit{XLNET} \cite{yang2020xlnet}, introduces the auto-regressive formulation to transformers and considers all permutations of the factorization order of the context while learning of bidirectional contexts.

A comparison of the performance of the studied models on the IMDB movie review dataset is reported under \textit{accuracy} in Table~\ref{tbl:exp2_onM}.

    \subsection{Explainability Models}
    Explainability entails various realizations such as visualization, attention interpretation and alignment \cite{zeilerF13,Chattopadhay_2018,sundararajan2017axiomatic,patro2019u} as surveyed in \cite{el2022survey}. We focus on \textit{rationale} extraction that supports a particular decision by highlighting causal input segments. Hence, we consider three explainability methods: Local Interpretable Model-agnostic Explanations (LIME) \cite{lime}, high-precision model-agnostic explanations (anchors) \cite{ribeiro2018anchors} and SHapley Additive exPlanations (SHAP) \cite{shap}. All these explainability models are black-box algorithms with a profound theoretical ground and a sound practical implementation.  

LIME \cite{lime} approximates \textit{any} model locally in a model-agnostic fashion. In NLP settings, LIME presents the interpretation in terms of bag-of-words regardless of what the NLP model originally accepts. This is achieved by approximating the classifier with a locally-more-explainable, potentially linear, model which is then trained on perturbed inputs in a local neighborhood.

Later, \cite{ribeiro2018anchors}  
define ``anchors'' as if-then rules to generate model agnostic explanations. These rules are a set of predicates, $A$, that are defined on interpretable representations in such a way that $A(x)$ returns $1$ if all its feature predicates are satisfied on $x$. Consequently, a rule is an anchor if $A(x) = 1$ and $A$ is a sufficient condition for the prediction $f(x)$. Then, random words, having the same Part-of-Speech (POS) TAG as ``absent'' tokens, are sampled and the ``absent'' tokens are then replaced by the random words to apply input perturbations and study their effect on the model prediction.  

Inspired by game theoretical concepts, \cite{shap} developed a unified approach to interpret model predictions. Analogous to the Shapley value computation, SHAP values are proposed as a unified measure that different explainability models, such as LIME \cite{lime}, Deep LIFT \cite{shrikumar2019learning} and layer-wise relevance propagation \cite{bach2015pixel}, approximate. SHAP is theoretically proven to satisfy \textit{local accuracy}, \textit{missingness} and \textit{consistency} as defined in \cite{shap}.

\section{Evaluation Dataset}\label{sec:dataset}
We anticipate the support of a group of ten data scientists who contributed to providing ground truth rationales. The labelers are between 21 and 26 years old balanced across gender. We consider the Rotten Tomatoes dataset \cite{data_rotten} with binary labels and we complement it with two additional labels: \textit{reasoning difficulty}, and \textit{extracted rationales}\footnote{Instructions to labelers are provided in the supplementary material}. 

First, the \textit{reasoning difficulty} reflects the complexity of reasoning about the sentiment and takes three values: $1, 2$, and $3$ with 3 being the most difficult. A special tag $4$ for \textit{reasoning difficulty} is used when a sentence requires additional context understanding. For example, the following review is labeled as 4: \textit{``not even the hanson brothers can save it''}. Explaining this sentence would require a higher-level knowledge of what the reviewer meant by \textit{hanson brothers} in a different context. It is noteworthy that the \textit{reasoning difficulty} does not reflect the readability nor the semantic complexity. It rather shows the difficulty of extracting evidence of a particular sentiment. 

Second, the \textit{extracted rationales} are a list of particular words in the sentence that contributed to the sentiment prediction. In curating the dataset, we alleviate human bias and subjectivity by asking more than one labeler to extract the rationales of each sentence. The aggregation of these explanations is done using the conjunction and disjunction operations. An additional correction step was applied to the aggregated explanations. 

Table~\ref{tbl:data_sample} shows samples from our dataset and Table~\ref{tbl:dataset} represents the dataset description aggregated by reasoning difficulty. Specifically, we have 1973 sentences with 20.9 words per sentence on average. Rationales consist of 3.1 words on average when merging (union) across labelers. When considering rationales that different labelers agree on (intersection), the sentences have 0.9 words on average. 934 sentences yield an empty set of rationale intersections. The latter statistic is important to show the degree of agreement between the labelers. A high number of empty intersections reflects diversity in the human thinking in the group of labelers.

\begin{table}[h]
\footnotesize
\begin{tabular}{p{6.3cm}C{1cm}C{1.2cm}C{1.8cm}C{1.8cm}}
\textbf{Sentence} & \textbf{Label} & \textbf{Reason-ing Diff.} & \textbf{Rationales (union)} & \textbf{Rationales (intersection)}\\ \hline \hline
beautiful to watch and holds a certain charm & 1 & 1 & beautiful charm & beautiful charm\\ \hline
it turns out to be smarter and more diabolical than you could have guessed at the beginning & 1 & 1 & smarter more diabolical & smarter diabolical\\ \hline
none of this is very original , and it isn't particularly funny. & 0 & 1 & none original is n't funny & none is n't funny \\ \hline
the film tunes into a grief that could lead a man across centuries. & 1 & 2 & grief & grief\\ \hline
the film is about the relationships rather than about the outcome . and it sees those relationships, including that between the son and his wife, and the wife and the father, and between the two brothers, with incredible subtlety and acumen. & 1 & 3 & incredible subtlety acumen & incredible subtlety \\ \hline
hard as this may be to believe, here on earth, a surprisingly similar teen drama, was a better film. & 0 & 3 & better & better\\ \hline
not even the hanson brothers can save it & 0 & 4 & - & - \\ \hline
more than simply a portrait of early extreme sports , this peek into the 1970s skateboard revolution is a skateboard film as social anthropology... & 1 & 4 & - & - \\ \hline \hline
\end{tabular}
\caption{Dataset sample. Rationales are individual words separated by a space.}\label{tbl:data_sample}
\end{table}

\begin{table}[h]
\centering
\begin{tabular}{lrrrrr}
Difficulty & 1 & 2 & 3 & 4 & \textbf{Total} \\
\hline\hline
Number of Sentences   &  1535 & 208 & 148 &  82 & \textbf{ 1973} \\
Words per sentence  &  20.1 & 24.2 & 23.7 & 23.2 & \textbf{ 20.9}\\
Words per explanation (union) &  3.4 & 2.9 & 2.6 & 0.0 & \textbf{ 3.1}  \\
Words per explanation (intersection)  &  1.07 & 0.7 & 0.45 & 0.0 & \textbf{0.9}\\

Number of Empty intersections & 629 &	121 &	102 & 82 & \textbf{934}\\
\hline\hline
\end{tabular}
\caption{Dataset description}\label{tbl:dataset}
\end{table}

\section{Methodology}\label{sec:method}
In what follows, we describe the design of our evaluation of faithfulness and plausibility and the formulation of our information retrieval-based metrics for \textit{plausibility} evaluation.

\subsection{Experimental Design}\label{sec:exp_setup}

Once the explanations are derived, two evaluation methods are adopted.
\subsubsection{\textit{Faithfulness}}
To evaluate the significance of the words in $E$ to the sentiment classification task objectively, we feed $E$ as an input to \mlmodel{} and we compute the prediction accuracy. This experiment does not require a comparison to ground-truth data; it rather studies the validity of the explanations through prediction. In this experiment, we investigate the effect of the ExAI model and its confidence on the explanation accuracy. 

\subsubsection{\textit{Plausibility}}
\begin{figure}
    \centering
    \includegraphics[width=0.6\textwidth]{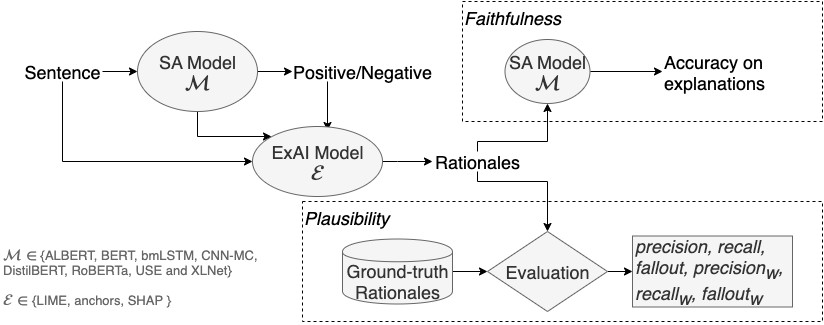}
    \caption{Experimental Design}
    \label{fig:exp_setup}
\end{figure}

After getting the explanations $E$ and their corresponding contributions $W$, a comparison with the ground truth labels $L$ allows us to compute our proposed metrics as in Section~\ref{sec:metrics}. 
Figure~\ref{fig:exp_setup} summarizes the experimental design.

\subsection{Plausibility Evaluation Metrics}\label{sec:metrics}
In what follows we assume the following: an explanation model \exmodel{} $\in$ $\{$ SHAP, anchors, LIME $\}$ is outputting a set of words $E = \{e_1, e_2, \dots, e_{N_1}\}$ with their corresponding contributions $W = \{w_1, w_2, \dots, w_{N_1}\}$  when explaining a machine learning model \mlmodel{}, where \mlmodel{} $\in \{$ ALBERT, BERT, bmLSTM, CNN-MC, DistilBERT, RoBERTa, USE, XLNet$\}$, on an input sentence $S = \{s_1, s_2, \dots, s_n\}$. To evaluate \textit{plausibility}, we compare $E$ to a sequence of ground truth words provided by human labelers $L = \{l_1, l_2, \dots,l_{N_2}\}$. Consequently, we propose three metrics inspired from information retrieval: \precision{}, \recall{} and \fallout{}.
The \precision{} computes the fraction of the words retrieved by \exmodel{} that are relevant from a human perspective. The \recall{} computes the fraction of words that are relevant from a human perspective and that are retrieved by \exmodel{}. The \fallout{} computes the fraction of non-relevant explanations that are retrieved from all the non-relevant words. 
\begin{align}
    \precision{} &= |L \cap E|/|E|\\
    \recall &= |L \cap E|/|L|\\
    \fallout &= | (S - L) \cap E|/|S - L|
\end{align}

where $|.|$, $\cap$, and $-$ are the set cardinality, intersection, and set difference respectively. 

A plausible rationale extraction is reflected in explanations that match human judgment, hence high precision and recall and low fallout rates. A high precision suggests that it is unlikely that the model will provide the word $s_i$ in the explanation if $s_i$ is not provided in the annotations. A high recall reflects that it is unlikely for the model to miss a word if it is in the annotations. We integrate the contribution scores $W$ to generate the weighted version of our metrics as follows:  
\begin{align}
    \precision{}_w &= |L \cap E|_W \hspace{0.5em}/\hspace{0.5em} |E|_W\\
    \recall{}_w &= |L \cap E|_W \hspace{0.5em}/\hspace{0.5em} |L|\\
    \fallout{}_w &= | (S - L) \cap E|_W \hspace{0.5em}/\hspace{0.5em}|S - L|
\end{align}

where $|.|_W$ is the weighted set cardinality computed as $|{s_1, s_2, s_M}|_W = w_1 + w_2 + w_M$, 
with $w_i \in W$ is the weight assigned to the word $s_i$ by the explainability method.

\section{Results and Discussion}\label{sec:results}
\subsection{Back-end Setup}
Before generating the explanations, the SA models were re-trained on the same dataset, IMDB movie reviews \cite{imdb}, according to their official code repositories. The obtained models are then explained by LIME, anchors, and SHAP on our provided dataset. All the experiments are run on Nvidia K80 GPU with 12 GB RAM and 0.82GHz memory clock rate. Following their official repositories, the considered back-end of the SA models varies between \textit{keras}, \textit{torch} and \textit{tensorflow}. Repositories implemented on Python 2.x were translated to Python 3.x and models were retrained accordingly.


\subsection{Evaluation of faithfulness}
Table \ref{tbl:exp2_onM} shows the initial model validation accuracy and its reduction when only the explanations are fed to the model \mlmodel{} with different confidences $\epsilon$. The parameters $\epsilon$ are tailored to the confidences that each explainer \exmodel{} produces. Anchors models produce explanations with confidence $\approx 1$ and thus no thresholds are considered. One can see that the explanations provided by LIME consistently outperform those provided by \textit{anchors} and SHAP on all SA models. This consistency is also maintained with different thresholds per model.  Predominantly, \textit{anchors} produces the least \textit{faithful} explanations yielding significant degradation of SA accuracy. 
Furthermore, the most interpretable models are predominantly transformers as bmLSTM, RoBERTa and ALBERT scored the highest according to LIME, anchors, and SHAP respectively. The consistency is not maintained on different SA models \mlmodel{}. For instance, bmLSTM is more explainable than BERT when LIME is used to explain but less explainable when SHAP is used.

More strikingly, LIME generally enhances the SA accuracy. This can suggest that the sentiment is highly concentrated in the rationales derived by LIME. SHAP, on the other, hand reduces this accuracy and the anchors model further significantly deteriorates it.
This can be explained by two factors. First, the models are all trained on full sentences; discontinuous chunks might mislead the model. Second, in some cases, the ExAI models fail to correctly label the negation in their provided explanations which might lead to misclassification. These results suggest that LIME's explanations are more absolute, hence useful, and can serve as standalone representations of the original sentences. 

\begin{table*}[h]
\begin{tabular}{lp{1.4cm}llllllll}

& & \multicolumn{3}{c}{\textbf{LIME}} &  \textbf{anchors} & \multicolumn{3}{c}{\textbf{SHAP}}  \\
\cmidrule{3-5} \cmidrule{7-10}\\
\textbf{Model}       & \textbf{Accuracy}                          & $\epsilon=0.1$       & $\epsilon=0.2$      & $\epsilon=0.3$      &  & $\epsilon=0.1$       & $\epsilon=0.2$      & $\epsilon=0.3$ & $\epsilon=0.5$ \\ \hline \hline
ALBERT & 89.55 & 93.15  & \underline{93.20}  & 93.16  & 68.30 & \textbf{88.54} & \textbf{88.34} & \textbf{88.54} & \textbf{88.54}\\
BERT & 89.86 & 92.60  & \underline{92.65}  & 92.44  & 67.28 & 87.63 & 87.53 & 87.20 & 86.56 \\
bmLSTM &  82.61 &	\textbf{\underline{93.93}} &	\textbf{\underline{93.93}} &	\textbf{93.83} &	77.38 &	76.84 &	76.28 &	75.87 &	75.22 \\
CNN-MC & 71.36 & 88.35  & 88.89  & \underline{88.90}  &  55.40 & 69.08 & 68.52 &	68.02 & 68.12  \\
DistilBERT & 88.77 & \underline{91.05}  & 90.84  & 90.70  & 63.91 & 88.40 & 88.10 & 88.35 & 87.97 \\
RoBERTa & \textbf{90.26}  & 88.34  & \underline{88.39}  & \underline{88.39}  & \textbf{82.32} & 88.25 & 88.24 & 88.21 & 88.19 \\
USE & 83.05 & 92.74 & 92.78 & \underline{92.87} & 80.43 &  75.45	 & 74.65 &	74.48 &	74.48  \\
XLNet & 90.11 & 92.61  & \underline{92.65}  & 92.60  & 71.50 & 87.83 & 86.87 & 86.87 & 85.85 \\

\hline\hline
\end{tabular}
\caption{Accuracy of \mlmodel{} on explanations of \exmodel{}. Highest accuracies per explanation model are highlighted in bold. Highest accuracies per SA model are underlined.}\label{tbl:exp2_onM}
\end{table*}

\subsection{Plausibility Evaluation on Ground Truth Data}

\begin{figure}[h]
    \centering
    \includegraphics[width=\textwidth]{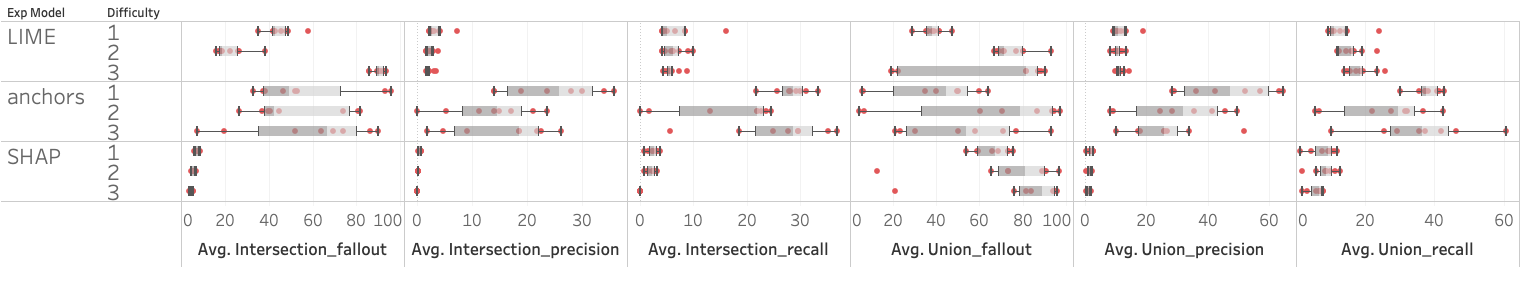}
    \caption{Explainability scores achieved by ExAIs (for different SA models)}
    \label{fig:dim_exai}
\end{figure}

Figure~\ref{fig:dim_exai} shows the distribution \precision{}, \recall{} and \fallout{} computed on the SA models explained by LIME, anchors and SHAP. Regardless of the reasoning difficulty, the anchors model consistently achieves higher scores on all models \mlmodel{} with the union and intersection operations. This observation can be explained by the fact that anchors' explanations are in harmony with human rationales, especially since anchors are defined as ``if-then'' rules. However, the anchors model exhibits higher variability and a consistent improvement is seen with the union operation giving the explanation a higher chance to be matched in the ground truth dataset. Finally, the reasoning difficulty level has an effect on the plausibility of explanations but the effect is not significant when aggregating across different SA and ExAI models.

As in the previous experiment, the explanation consistency is not maintained by \exmodel{}. For instance, the most explainable model according to anchors is XLNet followed by RoBERTa (as in the previous experiment), where CNN-MC and RoBERTa are the most explainable according to LIME and SHAP respectively. However, similar to the previous findings, the majority of transformer models disclose higher explainability scores compared to convolutional and recurrent ones. These results are consistent with the weighted metrics which further accentuate the outperformance of anchors' explanations over those of LIME and SHAP. Besides exhibiting better explainability, the anchors model is demonstrated to be more confident in generating its explanations.

\begin{figure}[h]
    \centering
    \includegraphics[width=\textwidth]{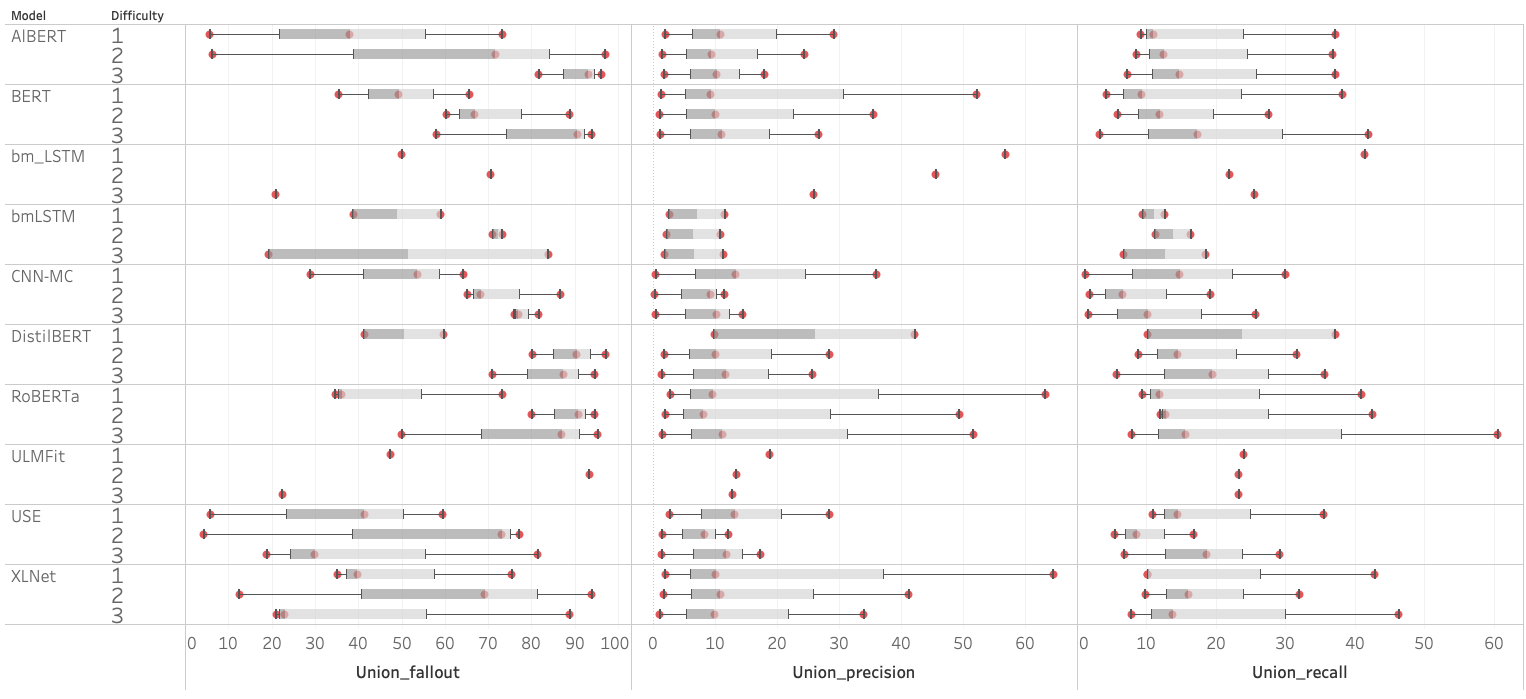}
    \caption{(Unweighted) explainability scores of SA models when varying the ExAI model}
    \label{fig:dim_model}
\end{figure}

Figure~\ref{fig:dim_model} shows the explainability scores achieved by the different SA models while varying \exmodel{}. Once again, transformers such as RoBERTa, BERT, and AlBERT achieve high agreements with ground-truth rationales. These results point to the crucial role of attention in matching human reasoning. Additionally, reasoning difficulty thwarts explainability, especially in non-transformer architectures such as CNN-MC and USE. 
 
\subsection{Qualitative Evaluation}
Inherently, instances with reasoning difficulty 4 are missing ground-truth explanations. We conduct a qualitative assessment of their explanations. For the sentence: \textit{``watching queen of the damned is like reading a research paper, with special effects tossed in''}, LIME produces \textit{tossed, paper, in, like, is, effects} and \textit{damned} as explanations with confidences $<0.3$ on RoBERTa. Clearly, the provided explanation is of low confidence and requires context knowledge such as boredom from reading a research paper. For sentences with a difficulty of 4, SHAP also provides an explanation with negligible weights (order of $10^{-3}-10^{-7}$).

\subsection{Take-Away Messages}
All dimensions have been considered, our evaluation allows us to conclude with the following speculations. LIME produces rationales with high ``sentiment concentration'' leading to more faithful explanations and \textit{useful} interpretations from a human perspective. The anchors model extracts supporting evidence that is more aligned with human judgment with higher confidence. On the level of the SA architecture, transformers can be, generally, deemed more faithful in deriving explanations. To a significant extent, attention can emulate human reasoning at the back of sentiment extraction. Nonetheless, no single tailor-made deep architecture is highly interpretable when changing the explainability lenses which motivates the ensembling of ExAI results. Finally, high confidence in the evidence extraction can be linked to reasoning simplicity rather than complexity. 



\section{Conclusion}\label{sec:conc}
In this work, we set the grounds for a rigorous evaluation framework for ExAI methods on SA models. We develop metrics to evaluate the explainability of SA models for their faithfulness and plausibility on our annotated rationales. We study the explainability of the SA methods from different angles: reasoning difficulty level, degree of agreement in the annotated data, ExAI model and its confidence level, and the underlying architecture of the SA model. Our empirical study derives important insights concerning the explainability of NLP models, SA specifically. Mainly the plausibility of anchors' explanations and faithfulness of LIME is demonstrated in our set of experiments. More importantly, we highlight the potential that attention mechanisms have in matching human reasoning. 

This work paves the way for a sound explainability evaluation framework that would engender user trust in NLP models and support behavioral marketing and market analysis systems. An immediate step in this line of work would be the extension to multi-target and multi-aspect sentiment analysis tasks and general NLP models. 

\section{Acknowledgment}
This work was supported by the University Research Board (URB)
and the Maroun Semaan Faculty of Engineering and Architecture (MSFEA) at the American University of Beirut.

\bibliographystyle{splncs04}
\bibliography{refs}


\end{document}